\documentclass[preprint,12pt]{elsarticle}
\usepackage{graphicx}
\usepackage{amssymb}
\usepackage{subfigure}
\usepackage{color}
\journal{Computer And Mathematics With Applications}

\begin{document}

\begin{frontmatter}

\title{Data complexity measured by principal graphs}


\author[curie,inserm,mines]{Andrei Zinovyev}
\author[leicester,railway]{Evgeny Mirkes}
\address[curie]{Institut Curie, rue d'Ulm 26, Paris, France, 75005}
\address[inserm]{INSERM U900, Paris, France}
\address[mines]{Mines ParisTech, Fontainebleau, France}
\address[leicester]{University of Leicester, University Road, Leicester, LE1 7RH}
\address[railway]{Krasnoyarsk Institute of Railway Engineering, Krasnoyarsk-28, 660028, Russia}

\begin{abstract}
How to measure the complexity of a finite set of vectors embedded in a multidimensional space? This is a non-trivial question which can be approached in many different ways. Here we suggest a set of data complexity measures using universal approximators, principal cubic complexes. Principal cubic complexes generalise the notion of principal manifolds for datasets with non-trivial topologies. The type of the principal cubic complex is determined by its dimension and a grammar of elementary graph transformations. The simplest grammar produces principal trees.

We introduce three natural types of data complexity: 1) geometric (deviation of the data's approximator from some ``idealized" configuration, such as deviation from harmonicity); 2) structural (how many elements of a principal graph are needed to approximate the data), and 3) construction complexity (how many applications of elementary graph transformations are needed to construct the principal object starting from the simplest one).

We compute these measures for several simulated and real-life data distributions and show them in the ``accuracy-complexity" plots, helping to optimize the accuracy/complexity ratio. We discuss various issues connected with measuring data complexity. Software for computing data complexity measures from principal cubic complexes is provided as well.

\end{abstract}


\end{frontmatter}


\section{Introduction}

\begin{flushright}
{\it To our scientific father, \\Prof. Alexander N. Gorban,\\ on his 60th birthday}
\end{flushright}

\label{}

\subsection{What is complex data?}

Rapid development of computer-based technologies in many areas of science, including physics, molecular biology, environmental research led to appearance of large datasets that are now characterized as ``Big Data" \cite{Lynch2008Big}. There is a tremendous challenge in how to store, analyze, query and visualize the Big Data. It is frequently said that the problem of the Big Data is not only that it is big but also that it is complex. Hence, it would be useful to define what ``complex data" means and be able to measure the complexity. This study is devoted to an attempt to define a way to measure some particular aspects of data complexity, connected to the data's geometry.

When somebody says ``I have complex data", this can mean many different things. This can refer to the number of measurements, heterogeneity of measurement types, variety of descriptor types, complexity of descriptors, impossibility or inability to formalize or abstract the data (like images), etc. Here we are going to deal with only one particular aspect of the complexity: complexity of data point distribution structure in some finite-dimensional space. We assume that a dataset can be represented as a set of vectors in a simple but potentially many-dimensional vectorial space. Formally, the question that we try to answer is: ``how complex is the finite distribution of vectors representing data points in $R^m$ space ($m > 1$), accompanied by some simple metrics"?

\subsection{Complexity of data as complexity of approximators}

There are many ways to approach the question formulated above. For example, describing ``gestalt" data clusters on the language of algebraic topology (persistent data homologies) can provide some insights into the complexity of the vector distribution's structure \cite{Zahn1971Graph, Zomorodian2005Topology}.  Akaike information criterion (AIC) can be used to select models of data of minimal complexity, using information theory \cite{Hirotugu1974new}.

Here we develop a different approach: we are going to substitute a distribution of data which potentially contains many points by a simpler object which will approximate the data (approximator). Then we will study the complexity of the approximator instead of the complexity of the data itself. By this we believe that our approximator is a good representation of the internal structure of the data, of the data's {\it gestalt}. A particular point distribution is an implementation of this gestalt, which can be characterized by bigger or smaller scattering of points from it, or by bigger or smaller number of data points. If two datasets have approximators of identical complexity then we postulate that the datasets have identical complexity too. This point of view implies that our measurements of data complexity should in general rather weakly depend on the number of points and the approximation error.

A good approximator always corresponds to a compromise between its accuracy and complexity. In classical data approximation methods, number of centroids in K-means, number of principal components, curvature or length of the principal curve serve as measures of complexity. A good approximator is able to catch the hypothetical intrinsic shape of the data distribution without trying to approximate the data's ``noise" (though ``one man's noise is another man's signal" \cite{Ng1990Quote}). Therefore, limiting approximator complexity is an important aspect of any data approximation strategy. Hence, if we provide 1) a measure of the approximation complexity and 2) a method to limit it, then we can define the complexity of a given data distribution as the complexity of the corresponding {\it optimal} approximator. As in the case of measuring effective intrinsic data dimension, there might exist a hierarchy of data complexity levels each corresponding to certain approximation ``depth".

Determining a trade-off between complexity and accuracy of approximation can be regarded as a particular application of the {\it Structural risk minimization principle} introduced by Vapnik and  Chervonenkis in 1974 \cite{VapnikChervonenkis1974}, if it is understood very generally (initially it was introduced for classification problems). Structural risk minimization principle is a model selection strategy which gives a model minimizing both empirical error and the model complexity (properly measured).

In practical applications, data approximation means a possibility of compressing the data. Less complex data are easier to store: a million points in thousand dimensional space can be simply distributed along a straight line with certain (relatively small) scattering around. Therefore, instead of storing the whole data massif, one can store the approximator structure, accompanied by some {\it uncertainty} estimates for various parts of it. This might be enough for any practical use of the data. This idea is by no means a new one (going back to the vector quantization \cite{Linde1980Algorithm} and the Minimum description length principle, from probability theory \cite{Rissanen1978}), but its concrete implementations and applications remain open questions: there is yet no standard ways and tools for compressing the vectorial data.

\subsection{Principal cubic complexes as universal approximators}

A fundamental problem on the way to implement the idea of looking at the data's complexity through its approximation consists in finding rather universal object able to approximate complex data structures and suggest a constructive algorithm to compute it. Gorban and Zinovyev in \cite{Gorban2007Topological} introduced a good candidate for this role: a principal cubic complex. Exact definition of it is given in the Methods section. In simple words, principal cubic complex is a Cartesian product of graphs (Figure~\ref{graphproduct}). One-dimensional principal cubic complex is simply a principal graph \cite{Gorban2009Handbook}. The graphs (factors) used to construct the cubic complex are produced by systematic application of some operations from a selected graph grammar. The operations can be scored according to how much they give in terms of optimization of some objective function (such as the elastic energy), and the best operation is applied to transform the graph.

The most trivial graph grammar consists in adding a node without connecting it to other graph nodes. This grammar produces the simplest possible ``approximator": a set of principal points \cite{Flury1990Principal, Gorban2009Handbook}. The well-known K-means clustering algorithm \cite{Steinhaus1956Sur} provides a way to estimate position of this set in dataspace. A bit more complex grammar allows to add a node to one of the terminal nodes of the graph (having only one or zero neighbours). This grammar produces one-dimensional grids which can represent curves. The Cartesian product of simple linear grids gives two-, three- and higher-dimensional grids able to represent hypersurfaces, embedded in the multidimensional space of data. When they approximate data in the sense of mean-squared error and satisfy some regularity (like smoothness) constraints, they are called ``principal curves" and ``principal manifolds" \cite{Hastie1984Principal, Gorban1999Neural, Kegl1999Principal, Gorban2005Elastic, Gorban2008Principal}. Putting requirement of linearity on these approximating surfaces corresponds to approximating data by lines and planes, known as Principal Component Analysis, invented by Pearson in 1901 \cite{Pearson1901lines}.

Next step in increasing the grammar complexity consists in allowing a node to be connected to any other node in the graph, or, be inserted in a middle of an existing edge. This produces an approximator called {\it the principal tree} which is already able to approximate various non-linear and branching data distributions. In this study we will use this first non-trivial case of graph grammar to estimate the complexity of some artificial and real data distributions.

There exists an infinite number of graph grammars able to produce more complex approximators (though we should define first the notion of the approximator's complexity). Hence, what we mean here by data complexity will drastically depend on the grammar choice: in other words, on how complex is the ``language" which we are going to use in order to describe the data.

\subsection{Three measures of the approximator's complexity}

We introduce three natural measures of the approximator's complexity, similar to the ones suggested in \cite{Gorban2009Handbook}: 1) Geometrical measure, 2) Structural measure, 3) Construction measure.

\subsubsection{Geometrical complexity}

The geometrical measure of complexity estimates the deviation of the approximating object from some ``idealized" configuration. The simplest such ideal configuration is linear:  in this case the nodes of the graph are located on some linear surface. Deviation from the linear configuration would mean some degree of {\it non-linearity}.

However, the notion of non-linearity is applicable only to relatively simple situations. For example, in the case of branching data distributions (see example in Figure~\ref{star_curve}), non-linearity is not applicable as a good measure of geometrical complexity. In \cite{Gorban2007Topological} it was suggested that a good generalisation of linearity as ``idealized" configuration can be the notion of {\it harmonicity}. An embedment of a graph into multidimensional space is called harmonic if, in each star of the graph, the position of the central node of the star coincides with the mean of its leaf vectors (see more formal definition in the Methods). A linear grid with equally spaced nodes is evidently harmonic. In order to deal with arbitrary graphs, representing various grids, one has to introduce the notion of pluriharmonicity, when the harmonicity is required only for a subset of stars or for some subsets of leaves in the stars (see Methods).

In our estimations of the geometrical complexity using principal trees we will use the deviation from a harmonic embedment as analogue and generalisation of the non-linearity.

\subsubsection{Structural complexity}

The structural complexity defines how complex is an approximator in terms of its structural elements (number of nodes, edges, stars of various degrees). In general, this index should be a non-decreasing function of these numbers. Contribution of some of the elements (for example, nodes and edges) might be not interesting for measuring the structural complexity and, hence, have zero weight (not present) in the resulting quantitative measure.

\subsubsection{Construction complexity}

We derive our approximators by the systematic application of the graph grammar operations, in the way which is the most optimal in terms of the objective function. One can introduce a measure of the approximator's complexity in the spirit of Kolmogorov (see, for example, \cite{Kolmogorov1965Three}). The complexity of a graph can be defined as a minimum number of elementary graph transformations which were needed to produce it, using given grammar. This measure can be similar to the structural complexity in some implementations but is not equivalent to it.

\subsection{Several remarks on measuring the data complexity}

Approximating data by complex objects (curves, trees) in our approach is connected to non-linear optimization with all its usual problems of existence of multiple local minima
and difficulties in finding the global minimum. We apply several tricks to better deal with these problems like the quadratic form of the energy functional to be minimized at each iteration, or gradual ``softening" of the grid \cite{Gorban2009Handbook}, but the problem can not completely disappear, of course. It is manifested already in the case of approximating data by principal points: K-means algorithm can converge to several local minima. Moreover, within the same accuracy, one can approximate the same dataset by different number of centroids. In practice multiple runs of K-means are needed to choose the most optimal configuration.

There are several exceptions here. One is approximating data by principal lines and linear manifolds. In this case, if the data does not have degeneracies in the covariation matrix, the quadratic energy functional has a unique global minimum. Another exception concerns a special class of data distributions which can be characterized as ``pseudo-linear": when the structure of their underlying ``gestalt" can be orthogonally mapped onto a line (see example in Figure~\ref{arc}). In other words, this corresponds to the situation when the data points can be naturally ordered using projection on a straight line. In this case, finding a close to global minimum is usually easy when starting from this line as an initial approximation.
Kernel PCA \cite{Schoelkopf1998Nonlinear} is another exception: it is able to produce non-linear approximating surfaces with a unique global optimum, but the approximation depends on the kernel form instead.

Non-uniqueness of the optimal approximator is tightly connected to the way of measuring the construction complexity. The graph grammar can contain the operations reducing the graph (for example, removing a node or an edge). The question is: should we take the actual (historical) number of graph grammar applications, or forget the learning history and count the number of steps which would be needed to produce the approximator's structure {\it de novo}? We leave this question open in this contribution because it appears as not very crucial in the case of principal trees that we use for estimating complexity.

Another remark concerns existence of clusters in data distributions. For the K-means approximator, the number $k$ is the only measure of approximator complexity. For more complex approximators, when the data is separated well into clusters, a relevant approach is to analyze the data complexity inside each cluster separately and then characterize their ``global" configuration. Real Big Data, however, is usually organized differently, with no sharp cluster borders and complicated (non-ellipsoidal) cluster shapes. In some cases, ``clusters" of data can even overlap but still represent two clearly different gestalts (like two overlapping circular distributions): a case which is very little addressed in the standard clustering methodology. For our purposes we will not separate the data distribution into clusters, assuming that the data distribution represents one relatively compact group of points. However, this question can be addressed by producing sets of unconnected principal objects, such as growing ``forest of principal trees" instead of growing one singular principal tree.

Our final remark concerns the dimensionality of data. Is the dimensionality itself an index of data complexity? The answer is not so simple. Of course, for the linear principal manifold, the number of retained components is a natural measure of the approximator's complexity. For more complex ones, first, we have to distinguish the dimensionality of the embedding space and the effective intrinsic dimensionality of data (dimensionality of the gestalt). Only the latter, of course, is in relation with data complexity. Second, higher dimension of data allows more complex patterns of data but not necessarily. Third, dimensionality of some objects (even as simple as principal trees) can be difficult to clearly define. And, moreover, data distributions are very frequently characterized by varying intrinsic dimension, being, for example, one-dimensional in some regions of data space and two- or three-dimensional in other regions (like the standard Iris dataset). We are not going to go deeply into these questions which have already been discussed in the literature \cite{Fukunaga1971algorithm, Kegl2002Intrinsic}.

\section{Materials and Methods}

\subsection{Elastic graphs and their (pluri)harmonicity}

In a series of works \cite{Gorban1999Neural, Gorban2001Vizualization, Gorban2005Elastic, Gorban2007Topological, Gorban2008Principal, Gorban2009Handbook, Gorban2010Principal} a metaphor of elastic membrane and plate was used to construct one-, two-
and three-dimensional principal manifold approximations of various
topologies. Mean squared distance approximation error combined with the
elastic energy of the membrane serves as a functional to be optimized. The
elastic map algorithm is extremely fast at the optimization step due to the
simplest form of the smoothness penalty. The methodology described below is based on these ideas.

Let $G$ be a simple undirected graph with set of vertices $V$ and set of edges $E$.
$k$-\textit{star} in a graph $G$ is a subgraph with $k+1$ vertices
$v_0,v_1,...,v_k \in V$ and $k$ edges \textbraceleft ($v_{0}, v_{i})$\textit{\textbar i }$=$
1$, .., k$\textbraceright$ \in E$. The \textit{rib} is by definition a 2-star.

Suppose that for each $k\ge$2, a family $S_{k}$ of
$k$-stars in $G$ has been selected. Then we define an \textit{elastic graph} as a graph with selected
families of $k$-stars $S_{k}$ and for which for all
$E^{(i)}\in E$ and $S_{k(j)} \in S_{k}$, the
corresponding elasticity moduli $\lambda_{i} \geq 0$ and $\mu_{kj} \geq 0$ are
defined.

\textit{Primitive} \textit{elastic graph} is an elastic graph in which every non-terminal node
(with the number of neighbours more than one) is associated with a $k$-star
formed by \textit{all} neighbours of the node. All $k$-stars in the primitive elastic graph
are selected, i.e. the $S_{k}$ sets are completely determined by the graph
structure.

Let $E^{(i)}(0)$, $E^{(i)}(1)$ denote two vertices of the graph edge
$E^{(i)}$ and \linebreak $S_{k(j)}(0),...,S_{k(j)}(k)$ denote vertices of
a $k$-star $S_{k(j)}$ (where $S_{k(j)}(0)$ is the central vertex, to
which all other vertices are connected). Let us consider a map $\varphi
:V~\to \textbf{R}^{m}$ which describes an embedding of the graph into a
multidimensional space. The \textit{elastic energy of the graph embedding in the Euclidean space }is defined as

\begin{equation}
\label{elastic_energy}
U^{\varphi}(G):=U_{E}^\varphi(G)+U_R^\varphi(G),
\end{equation}
\begin{equation}
\label{elastic_energy_stretching}
U_{E}^{\varphi }(G):=\sum\limits_{E^{(i)}} {\lambda_{i} \left\| {\varphi
(E^{(i)}(0))-\varphi (E^{(i)}(1))} \right\|^{2}} ,
\end{equation}
\begin{equation}
\label{elastic_energy_bending}
U_{R}^{\varphi}(G):=\sum\limits_{S_{k}^{(j)} } {\mu_{kj} \vert \vert
\varphi (S_{k}^{(j)} (0))-\frac{1}{k}\sum\limits_{i=1}^k {\varphi
(S_{k}^{(j)} (i))} \vert \vert^{2}} .
\end{equation}

Let us make two remarks. The values $\lambda_{i}$ and $\mu_{kj}$ are the
coefficients of stretching elasticity of every edge $E^{(i)}$ and
of resistance to harmonicity violation for every star $S_{k}^{(j)}$ (which is an analogue of star ``rigidity").
In the simplest case $\lambda_{1}$~$=$~$\lambda_{2}$~$=$~$...$~$=$~$\lambda
_{s}~=$~$\lambda (s), \mu_{k1}= \mu_{k2} = ... = \mu_{kr} =
\mu(r)$, where $s$ and $r$ are the numbers of edges and stars correspondingly.

$U_{E}^{\varphi }(G)$ penalizes the total length of the edges and thus
provides regularization of distances between node positions.
After the graph embedding is computed, $\lambda_{i}$
can be put to zero with little effect on the graph configuration.

A map $\varphi:V\to \textbf{R}^{m}$ defined on vertices of $G$ is
\textit{pluriharmonic} iff for any $k$-star $S_{k}^{(j)} \in S_{k}$ with the central vertex
$S_{k}^{(j)}(0)$ and the neighbouring vertices $S_{k}^{(j)}(i)$,
$i=1...k$, the equality holds:
\begin{equation}
\label{eq1}
\varphi(S_{k}^{(j)} (0))=\frac{1}{k}\sum\limits_{i=1}^k {\varphi
(S_{k}^{(j)}(i))}.
\end{equation}

Pluriharmonic embedding for primitive graphs is called simply harmonic embedment.
For a perfectly pluriharmonic embedding the last component $U_{R}^{\varphi}(G)$
of the elastic energy is zero, i.e. minimal.

\subsection{Elastic principal graph with given structure}

Let us choose an elastic graph, characterized by some structure.
\textit{Elastic principal graph} for a dataset $X$ is an elastic graph
embedded in $\textbf{R}^{m}$ using such a map $\varphi
_{opt}:V\to \textbf{R}^{m}$ that corresponds to the minimal value of
the functional

\begin{equation}
\label{total_energy}
U^{\varphi }(X,G)=MSD(X,G)+U^{\varphi}(G),
\end{equation}

\noindent where the mean squared distance (MSD) from the dataset $X$ to the elastic
graph $G$ is calculated as the distance to the finite set of vertices
\textbraceleft \textbf{y}$^{1}=\varphi (v_{1})$,...,
\textbf{y}$^{k}=\varphi(v_{k})$\textbraceright, i.e. for each datapoint
the closest node of the graph is determined and the MSD is the mean of all such
squared distances.

\subsection{Definition of the elastic principal cubic complex}

\begin{figure}\center{
        \includegraphics[width=8cm]{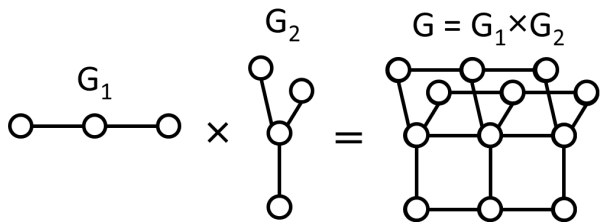}
\caption{\label{graphproduct} Simple example of a Cartesian product of two factor-graphs.}}
\end{figure}

Introducing principal cubic complexes \cite{Gorban2007Topological} gives a way to use ``$r$-dimensional" graphs
(similar to $r$-dimensional manifolds, see Figure~\ref{graphproduct}).

\textit{Elastic cubic complex} $K$ \textit{of intrinsic dimension} $r$ is a Cartesian product
$G_{1}\times ... \times G_{r}$ of elastic graphs $G_{1},...,G_{r}$. We call each $G_{i}$ a factor of the cubic complex.
Each factor is composed of a vertex set $V_{i}$: hence, the set of vertices $V$ of the cubic complex is $V = V_{1}\times ... \times V_{r}$.

Let us select a factor $G_{i}$, $i\in 1...r$ and any node in another factor $v_{j} \in V_{j}(j\ne i)$.
For a chosen set of vertices $V_{i}$ and a node $v_{j} \in V_{j}(j\ne i)$, there is a copy of $G_{i}$ in $G$.
It is defined by

1) vertices $$(v_{1},...,v_{i-1},v,v_{i+1},...,v_{r}) (v\in V_{i}),$$

2) edges
$$((v_{1},...,v_{i-1}, v, v_{i+1},...,v_{r}), (v_{1},...,v_{i-1}, v', v_{i+1},...,v_{r})), (v,v')
\in E_{i},$$

3) $k$-stars in the form $$(v_{1},...,v_{i-1}, S_{k}, v_{i+1},...,v_{r}),$$
where $S_{k}$ is a $k$-star in $G_{i}$.

For any $G_{i}$ there are
$\prod\limits_{j\ne i} {\vert V_{j} \vert }$ copies of $G_{i}$ in $G$. Sets of
edges and $k$-stars for Cartesian product are unions of that set through all
copies of all factors. A map $\varphi:V_{1}\times ... \times V_{r} \to \textbf{R}^{m}$ maps
all the copies of factors into $\textbf{R}^{m}$ too.

By construction, \textit{the energy of the elastic graph product is the energy sum of all factor copies. }It is, of course, a quadratic functional
of $\varphi$.

If we approximate multidimensional data by a $r$-dimensional object, the number
of points (or, more generally, elements) in this object grows with $r$ exponentially.
This is an obstacle for grammar--based algorithms even for modest $r$, because
for analysis of the rule $A \to B$ applications we should investigate all
isomorphic copies of $A$ in $G$. Introduction of a cubic complex is useful
factorization of the principal object which allows to avoid this problem.

\subsection{Basic algorithm for optimization of the elastic principal graph with given structure}

In the Euclidean space one can apply an Expectation-Maximization (EM) algorithm for computing the optimal embedding
map $\varphi$ of an elastic principal graph for a finite dataset $X$.

\begin{enumerate}
\item Choose some initial position of nodes of the elastic graph \textbraceleft $\textbf{y}^{1}=\varphi (v_{1}),...,\textbf{y}^{k}=\varphi(v_{k})$\textbraceright, where $k$ is the number of graph nodes $k=$ \textbar $V$\textbar ;
\item Calculate two matrices $e_{ij}$, and $s_{ij}$, using the following sub-algorithm:
\begin{enumerate}
\item Initialize the $s_{ij}$ matrix to zero;
\item For each $k$-star $S_{k}^{(i)}$ with elasticity module $\mu_{ki}$, outer nodes $v_{N1},~...~,v_{Nk}$ and the central node $v_{N0}$, the $s_{ij}$ matrix is updated as follows $(1\le l,m \le k)$:
\[{\begin{array}{*{20}c}
 {s_{N_{0} N_{0} } \leftarrow s_{N_{0} N_{0} } +\mu_{ki} ,\;\;s_{N_{l}
N_{m} } \leftarrow s_{N_{l} N_{m} } +{\mu_{ki} } \mathord{\left/ {\vphantom
{{\mu_{ki} } {k^{2}}}} \right. \kern-\nulldelimiterspace} {k^{2}}} \hfill
\\{s_{N_{0} N_{l} } \leftarrow s_{N_{0} N_{l} } -{\mu_{ki} } \mathord{\left/
{\vphantom {{\mu_{ki} } k}} \right. \kern-\nulldelimiterspace}
k,\;\;s_{N_{l} N_{0} } \leftarrow s_{N_{l} N_{0} } -{\mu_{ki} }
\mathord{\left/ {\vphantom {{\mu_{ki} } k}} \right.
\kern-\nulldelimiterspace} k} \hfill \\
\end{array} }\]
\item Initialize the $e_{ij}$ matrix to zero;
\item For each edge $E^{(i)}$ with weight $\lambda_{i}$, one vertex $v_{k1}$ and the other vertex $v_{k2}$, the $e_{jk}$ matrix is updated as follows:
\[{\begin{array}{*{20}c}
 {e_{k_{1} k_{1} } \leftarrow e_{k_{1} k_{1} } +\lambda_{i} ,\;\;e_{k_{2}
k_{2} } \leftarrow e_{k_{2} k_{2} } +\lambda_{i} } \hfill \\
 {e_{k_{1} k_{2} } \leftarrow e_{k_{1} k_{2} } -\lambda_{i} ,\;\;e_{k_{2}
k_{1} } \leftarrow e_{k_{2} k_{1} } -\lambda_{i} } \hfill \\
\end{array} }\]
\end{enumerate}
\item Partition $X$ into subsets $K_{i}$, $i=$1..$k$ of data points by their proximity to $\textbf{y}_{k}:K_{i}=\{{\rm {\bf x}\in X}:{\rm {\bf y}_i} =\arg\min\limits_{{\rm {\bf y}_j} \in Y} \mbox{dist}({\rm {\bf x}},{\rm {\bf y}_j})\}$;
\item Given $K_{i}$ , calculate matrix $a_{js} =\frac{|K_j| \delta_{js} }{|X| }+e_{js} +s_{js}$, where $\delta_{js}$ is the Kronecker's symbol.
\item Find new position of~\textbraceleft $\textbf{y}^{1},...,\textbf{y}^{k}$\textbraceright~by solving the system of linear equations
\[\sum\limits_{s=1}^k {a_{js} {\rm {\bf y}}^{s}} =\frac{1}{|X| }\cdot \sum\limits_{{\rm {\bf x}}^{i}\in K_{j} } { {\rm {\bf
x}}^{i}}\]
\item Repeat steps 3-5 until complete or approximate convergence of node positions \textbraceleft $\textbf{y}^{1},...,\textbf{y}^{k}$\textbraceright.
\end{enumerate}

As usual, the EM algorithm described above gives only locally optimal
solution. One can expect that the number of local minima of the energy
function $U$ grows with increasing the `softness' of the elastic graph
(decreasing $\mu_{kj}$ parameters). Because of this, in order to obtain a
solution closer to the global optimum, the \textit{softening strategy} has been proposed, used in the
algorithm for estimating the elastic principal manifold \cite{Gorban2009Handbook}.

\subsection{Graph grammars}

The graph grammars \cite{Nagl1976Formal} provide a well-developed formalism for the description of elementary transformations. An elastic graph grammar is presented as a set of production (or substitution) rules. Each rule has a form $A\leftarrow B$, where A and B are elastic graphs. When this rule is applied to an elastic graph, a copy of $A$ is removed from the graph together with all its incident edges and is replaced with a copy of $B$ with edges that connect $B$ to the graph.

Let us define \textit{graph grammar} $O$ as a set of graph grammar operations $O=$\textbraceleft
$o_{1},..,o_{s}$\textbraceright. All possible applications of a graph grammar
operation $o_{i}$ to a graph $G$ gives a set of transformations of the initial
graph $o_{i}(G)=$ \textbraceleft $G_{1}$, $G_{2}$,...,
$G_{p}$\textbraceright, where $p$ is the number of all possible applications of
$o_{i}$ to $G$. Let us also define a sequence of $r$ different graph grammars~
$\{O^{(1)}=\{o_{1}^{(1)} ,...,o_{s_{1} }^{(1)} \}\;,\;\cdots
,O^{(r)}=\{o_{1}^{(r)} ,...,o_{s_{r} }^{(r)} \}\}$.

Let us choose a grammar of elementary transformations, predefined boundaries
of structural complexity $\textit{SC}_{max}$, construction complexity $\textit{CC}_{max}$ and elasticity coefficients $\lambda_{i}$ and $\mu_{kj}$ .

Using these ingredients, we can choose the structure of the elastic principal graphs among all possible
graph structures that can be obtained by application of the given graph grammar.

\textit{Elastic principal graph} for a dataset $X$ is such an elastic graph
$G$ embedded in the Euclidean space by the map $\varphi$:$V\to
~$\textbf{R}$^{m\, \, }$that SC($G)\le $ \textit{SC}$_{max\, }$, CC($G)\le $\textit{ CC}$_{max\, }$,
and $U^{\varphi }(X,G)$$\to$~min over all possible elastic graphs $G$ embeddings
in \textbf{R}$^{m\, }$.

Note that this definition does not define a unique elastic principal graph: for the same dataset, one can have several
principal graphs with equal $U^{\varphi }(X,G)$.

\subsection{Algorithm for the principal graph construction}

\begin{enumerate}
\item Initialize the elastic graph $G$ by 2 vertices $v_{1}$ and $v_{2}$ connected by an edge. The initial map $\varphi$ is chosen in such a way that $\varphi (v_{1})_{\, }$ and $\varphi (v_{2})$ belong to the first principal line in such a way that all the data points are projected onto the principal line segment defined by $\varphi (v_{1})$, $\varphi (v_{2})$;
\item For all $j=$1\textellipsis $r$ repeat steps 3-6:
\item Apply all grammar operations from $O^{(j)}$ to $G$ in all possible ways; this gives a collection of candidate graph transformations \textbraceleft $G_{1}$, $G_{2}$, \textellipsis \textbraceright ;
\item Separate \textbraceleft $G_{1}$, $G_{2}$, \textellipsis \textbraceright ~into  \textit{permissible} and \textit{forbidden} transformations; permissible transformation $G_{k}$ is such that $SC(G_{k})\le SC_{max}$, where $SC_{max}$ is some predefined structural complexity upper bound;
\item Optimize the embedment $\varphi$ and calculate the energy functional $U^{\varphi}(X,G)$ of the graph embedment for every permissible candidate transformation {\it after optimisation}, and choose such a graph $G_{opt}$ that gives the minimal value of the energy functional: $G_{opt} =\arg\min\limits_{G_{k}\in permissible\;set} U^{\varphi } (X,G_{k} )$;
\item Substitute $G\leftarrow G_{opt}$;
\item Repeat steps 2-6 until the set of permissible transformations is empty or the number of operations exceeds a predefined number -- the construction complexity.
\end{enumerate}

\subsection{Principal trees}

\textit{Principal tree} is an acyclic primitive elastic principal graph.

\textit{`Add a node, bisect an edge'} graph grammar $O^{(grow)}$ applicable for the class of
primitive elastic graphs consists of two operations: 1) The transformation
\textit{``add a node'' }can be applied to any vertex $v$ of $G$: add a new node $z$ and a new edge ($v,~z)$; 2) The
transformation \textit{``bisect an edge'' }is applicable to any pair of graph vertices $v, v'$ connected by an
edge ($v, v')$: delete edge ($v, v')$, add a vertex $z$ and two edges, ($v, z)$ and ($z, v')$. The
transformation of the elastic structure (change in the star list) is induced
by the change of the tree's edges, because the elastic graph is primitive.
Consecutive application of the operations from this grammar generates trees,
i.e. graphs without cycles.

Application of the {\it ``add a node"} graph grammar operation should be accompanied by a concrete recipe on how to define the map $\varphi(z)$ (position in the data space) for a new node $z$. When there is only one node in the graph, there is no evident strategy on how to do this. In practice we start constructing the graph from two nodes connected by an edge, positioned on the first principal component. Another solution is to use a random node positioning: however, most probably the first iteration will orient it close to the first principal component.

If the graph contains two or more vertices then the harmonicity is used to define the map $\varphi(z)$. For the transformation {\it ``add a node"}, the newly formed $k$-star containing the new node $z$ has to be pluriharmonic, and the formula~(\ref{eq1}) is used to compute the map $\varphi(z)$. For the ``bisect an edge" graph grammar operation, the newly formed rib (a 2-star) with the central vertex $z$ should be pluriharmonic. In this case, the formula to define the map $\varphi(z)$ is very simple: $\varphi(z) = \frac{1}{2}(\varphi(v)+\varphi(v'))$.

\textit{`Remove a leaf}, \textit{remove an edge'} graph grammar $O^{(shrink)}$ applicable for
the class of primitive elastic graphs consists of two operations: 1) The
transformation `\textit{remove a leaf}' can be applied to any vertex $v$ of $G$ with connectivity degree
equal to 1: remove $v$ and remove the edge ($v$,$v'$) connecting v to the tree; 2)
The transformation `\textit{remove an edge}' is applicable to any pair of graph vertices $v, v'$ connected
by an edge ($v, v')$: delete edge ($v, v')$, delete vertex $v'$, merge the $k$-stars for
which $v$ and $v'$ are the central nodes and make a new $k$-star for which $v$ is the
central node with a set of neighbours which is the union of the neighbours
from the $k$-stars of $v$ and $v'$.

Also we should define the structural complexity measure \linebreak
SC($G)=$SC(\textbar $V$\textbar ,\textbar $E$\textbar ,\textbar $S_{2}$\textbar
,\textellipsis ,\textbar $S_{m}$\textbar ). Its concrete form depends on the
application field. Here are some simple examples:

\begin{enumerate}
\item SC($G)=$ \textbar $V$\textbar : i.e., the graph is considered more complex if it has more vertices;
\item $\mbox{SC(}G\mbox{)\, =\, }\left\{ {{\begin{array}{*{20}c}
 {\vert S_{3} \vert ,\;\mbox{if}\;\vert S_{3} \vert \le b_{\max } \;\mbox{and}\;\sum\limits_{k=4}^m {\left| {S_{k} } \right|} =0} \hfill \\
 {\infty ,\;\;\mbox{otherwise}} \hfill \\
\end{array} }} \right.,$
\end{enumerate}
i.e., only $b_{max}$ simple branches (3-stars) are allowed in the principal
tree.

Using the sequence \textbraceleft $O^{(grow)}$,$ O^{(grow)}$,$ O^{(shrink)}$\textbraceright
~in the above-described algorithm for estimating the elastic principal graph
gives an approximation to the principal trees. Introducing the `tree
trimming' grammar $O^{(shrink)\, }$allows to produce principal
trees closer to the global optimum, trimming excessive tree branching and
fusing \textit{k{}}-stars separated by small `bridges'.

\subsection{Complexity measures used in the examples}

In the examples below, we used principal trees constructed for several artificial and real-life data distributions, using the following forms of the complexity measures.

For measuring the geometrical complexity, we used the last term in the energy function (\ref{elastic_energy_bending}), which penalizes deviation from the harmonic tree shape. For measuring complexity here, we put all $\mu_{kj}=1$. We found out that it is convenient to multiply this term by the number of nodes squared, i.e. we used the following form of the geometrical complexity $GC$ of graph $G$ embedded in the multidimensional space by the map $\varphi$:

\begin{equation}
\label{geometrical_complexity}
GC^{\varphi}(G) = N_{nodes}^2U_{R}^{\varphi}(G).
\end{equation}

Using the $N_{nodes}^2$ multiplier makes $GC$ closer to the sum of squared second derivative discrete estimations for the ribs and its analogue for the stars, i.e. $\frac{\sum\limits_{i=1}^k {\varphi
(S_{k}^{(j)} (i))-k\varphi(S_{k}^{(j)}(0))}}{(\frac{1}{k}\sum_{i=1}^k{|\varphi (S_{k}^{(j)}(i))-\varphi (S_{k}^{(j)}(0))|})^2}$. Adding new nodes when bisecting edges results in decreasing the average length of the graph's edges, and the $N_{nodes}^2$ multiplier compensates this effect. We have checked numerically the correct scaling of (\ref{geometrical_complexity}) with the increasing number of nodes for several typical growing graphs.

For the structural complexity, below we do not introduce any quantitative measure, but use a symbolic barcoding for showing the number of structural graph elements (nodes, 3-stars, 4-stars, etc.):

\begin{equation}
\label{structural_complexity}
SC(G) = N_{k-stars}| ... | N_{4-stars} | N_{3-stars} || N_{nodes}.
\end{equation}

For example, ``$2|6||15$" means a principal tree with 15 nodes, 6 stars of order 3 and 2 stars of order 4. We do not show the number of edges and ribs in the barcode because the number of edges in the tree is always $N_{nodes}-1$. The number of ribs also can be easily computed from the number of nodes and number of $k$-stars ($k>2$), and it does not characterize the tree topology, but rather the number of nodes inserted in the tree branches.

The construction complexity of the principal trees which are produced by only applying grammar operations adding one node at a time, equals, evidently, $N_{nodes}-1$, which makes it a particular case of the structural complexity. This is also true if only the final structure of the principal tree is analyzed forgetting the historical sequence of graph grammar applications. Of course, the construction complexity can be different from the structural complexity in the case of less simple graph grammars. Nevertheless, one can imagine a scenario when the {\it historical} construction complexity is not trivial for principal trees also. For example, this might be achieved if no trimming operation is applied when the increase of the elastic energy is too big. Then a sequence of graph grammar applications can contain any number of growing and trimming operations (provided that the first is bigger than the second, of course), and the resulting historical construction complexity does not equal $N_{nodes}-1$. Having all this in mind, we nevertheless do not use the construction complexity explicitly in the examples below.

\subsection{Available implementations}

Method for constructing elastic principal graphs (including principal curve, principal manifold and principal tree) is implemented in Java language.
User-friendly graphical interface for constructing principal manifolds is available at http://bioinfo.curie.fr/projects/vidaexpert.
User-friendly graphical interface (Java-applets) for constructing principal trees in 2D is available at http://bioinfo.curie.fr/projects/elmap. The software found applications
in microarray data analysis, visualization of genetic texts, visualization
of economical and sociological data and other fields \cite{Gorban2001Vizualization, Gorban2003Application, Gorban2005Elastic, Gorban2008Principal, Gorban2010Principal}.

\section{Results and Discussion}

\subsection{Test examples}

\begin{figure}
 \subfigure
    {
        (a)~~\includegraphics[width=11.5cm]{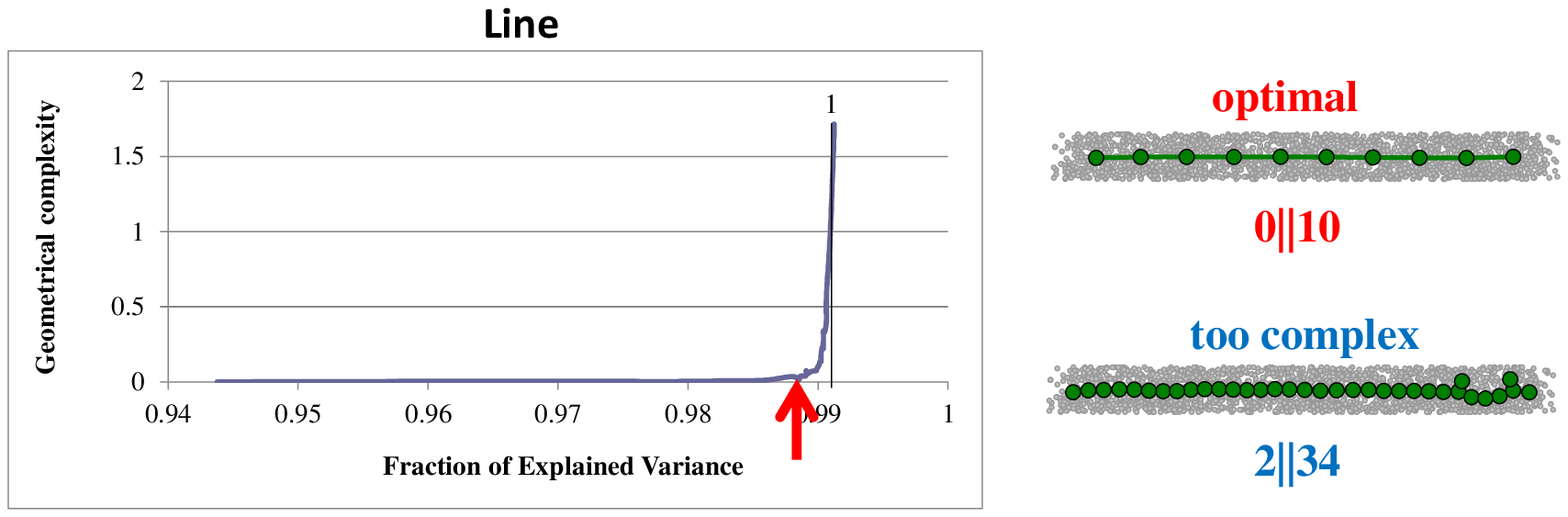}
        \label{linear}
    }
 \subfigure
    {
        (b)~~\includegraphics[width=11.5cm]{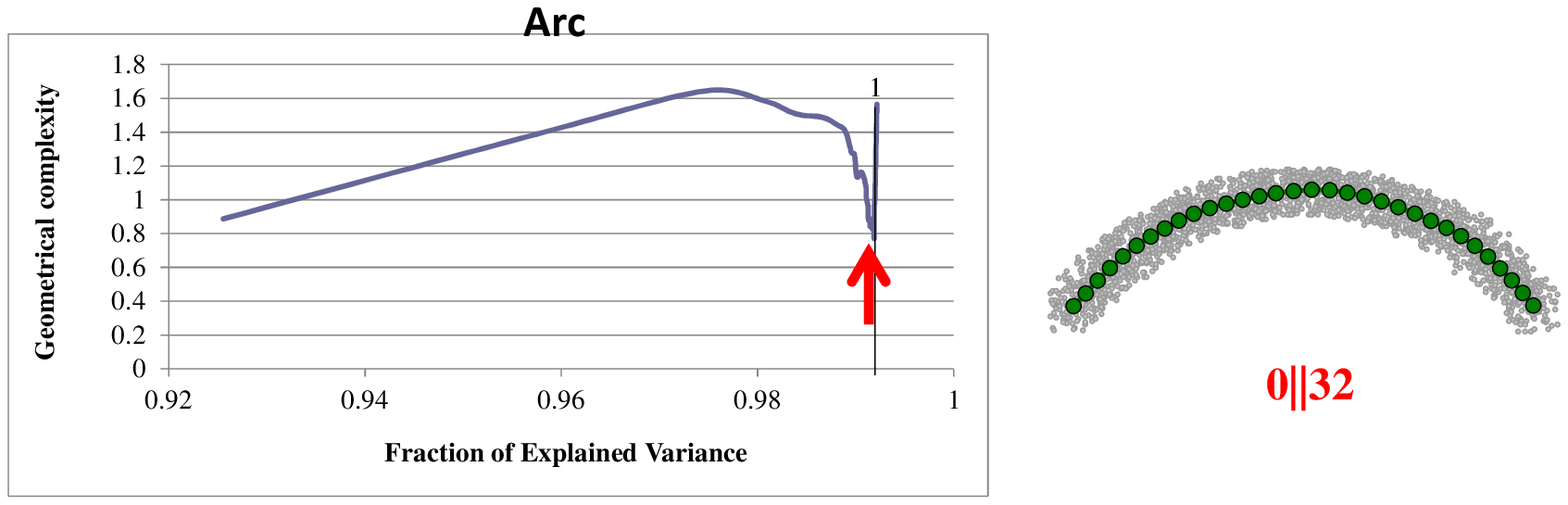}
        \label{arc}
    }
  \subfigure
    {
        (c)~~\includegraphics[width=11.5cm]{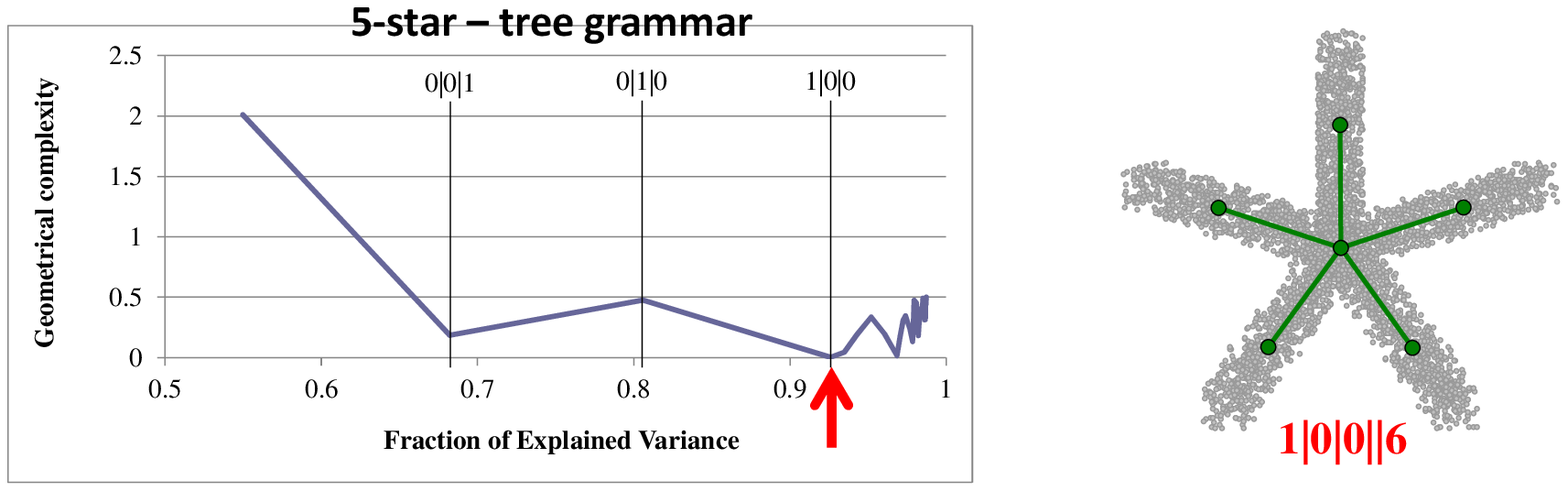}
        \label{star_tree}
    }
 \subfigure
    {
        \hspace{1mm}(d)~~\includegraphics[width=11.5cm]{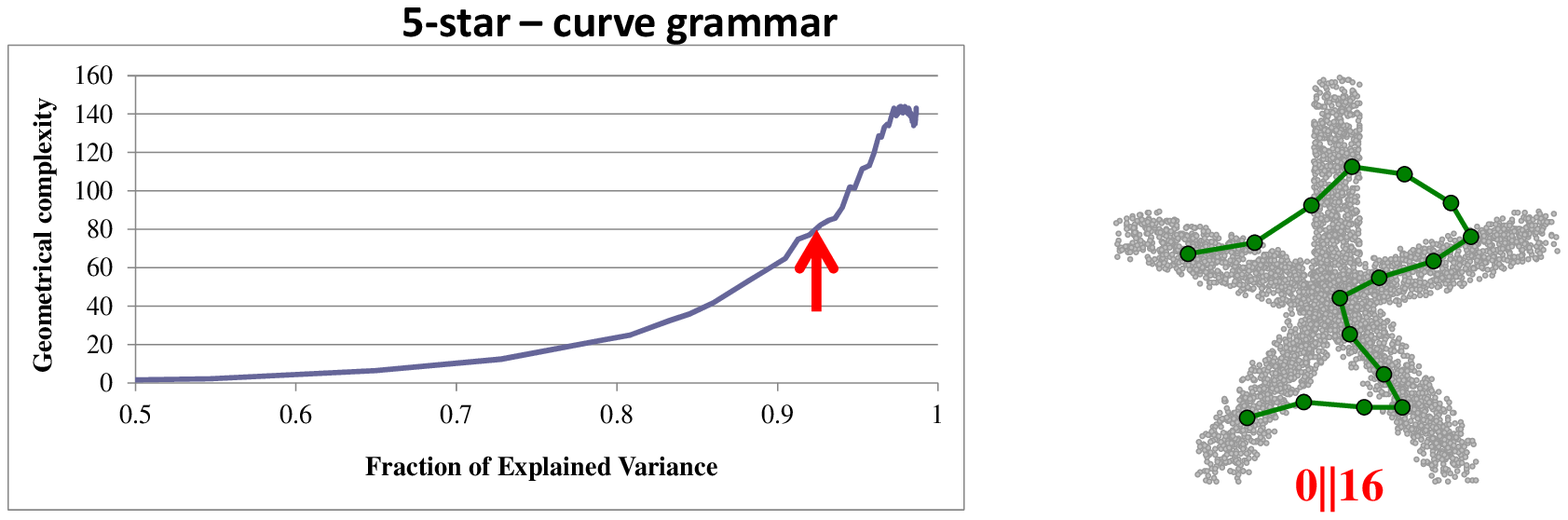}
        \label{star_curve}
    }
\caption{\label{test_examples} Examples of the ``accuracy-complexity" graphs (on the left) for the test 2D data distributions. The arrows show the accuracy/complexity chosen to construct the principal tree or curve shown on the right together with the data. The barcode shown with the principal tree or curve specifies the structural complexity. Part of the barcode (only the number of stars) is shown on the ``accuracy-complexity" plot above the vertical lines visualizing the discrete changes of the structural complexity. }
\end{figure}

Let us first introduce the ``accuracy-complexity" plots that we will use to find the optimally complex data's approximator (see examples in Figure~\ref{test_examples}). On the abscissa of the plot we show the Fraction of Variance Explained (FVE), i.e. a unity minus ratio between the Mean Squared Error and the total data variance. The Mean Squared Error is measured with respect to the closest distance to the approximator as a polyline, i.e. to its closest node or the closest edge. On the ordinate of the plot we show the geometrical complexity defined by the formula (\ref{geometrical_complexity}). The absolute value of the geometrical complexity is in general not comparable between datasets (because its scale changes with the intrinsic data dimensionality and the spatial data scale). What is informative in the plot is the structure of minima and maxima of the geometrical complexity as well as its behavior when the approximator approaches 100\% of explained variance. The changes in structural complexity are shown in the plot by vertical lines labeled by the barcode defined in (\ref{structural_complexity}).

\begin{figure}
        \includegraphics[width=13cm]{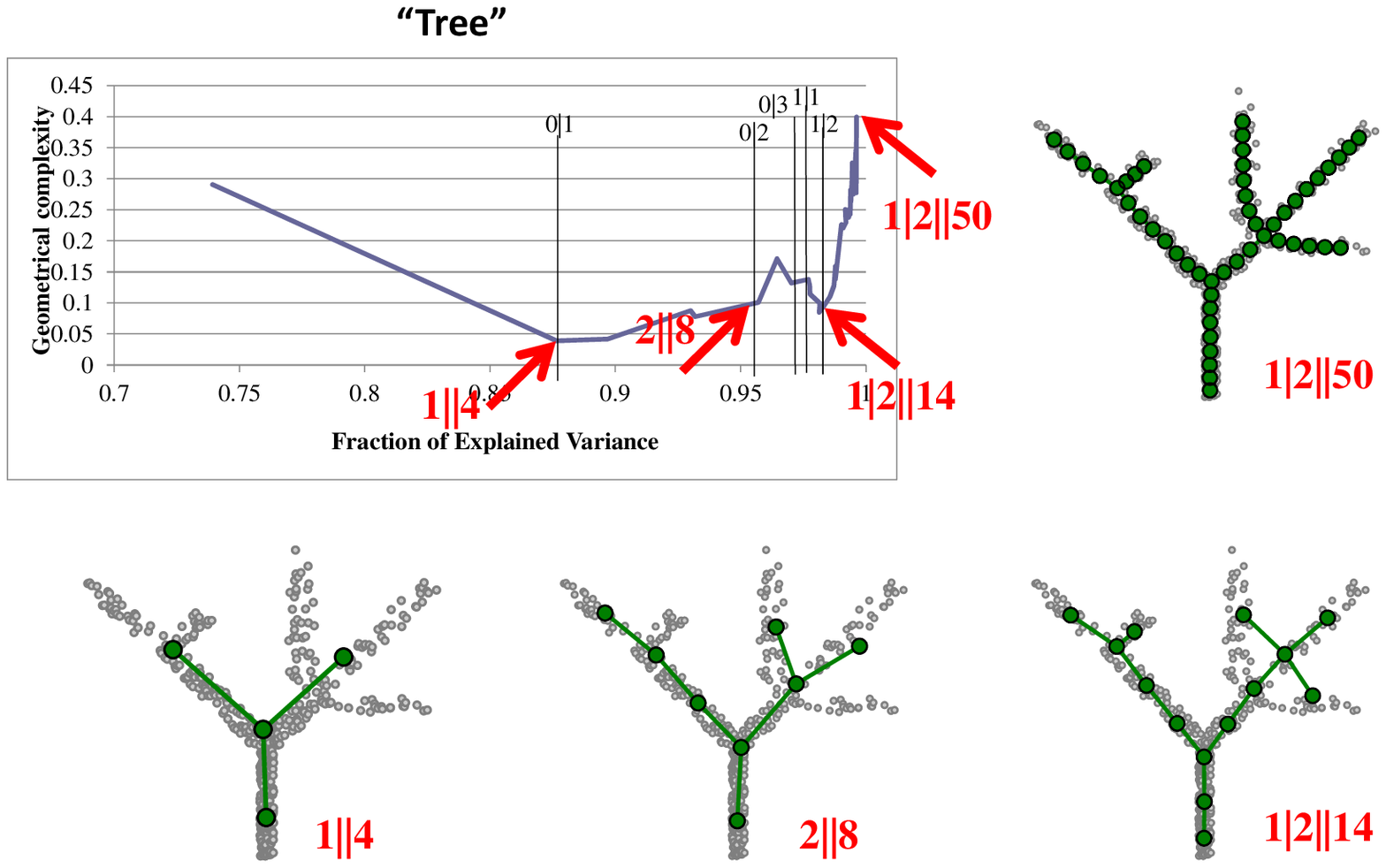}
        \label{tree}
\caption{\label{tree_example} The ``accuracy-complexity" graph for a tree-like 2D data distribution. Several scales of the approximator's complexity are shown. Two of them, corresponding to the structural complexity barcodes $1||4$ and $1|2||14$ are optimal and approximate the distribution structure at a certain ``depth".}
\end{figure}

To calibrate and understand the behavior of the ``accuracy-complexity" graph, we used several simple 2D distributions (Figures~\ref{test_examples},\ref{tree_example}). For example, a simple linear distribution shown in Figure~\ref{linear} leads to a very simple ``accuracy-complexity" plot. The geometrical complexity remains close to zero but drastically grows up close to $FVE \approx 0.99$. At some point, the approximator starts to produce branches which, evidently, approximate some noisy local data structures and, hence, correspond to excessive complexity. In Figure~\ref{linear},right we show the optimal (corresponding to a not very deep local minimum) principal tree containing 10 nodes, no branching, and the principal tree obtained at 34 nodes, containing two 3-stars (an example of an approximator whose structure is too complex with respect to the data distribution).

Figure~\ref{arc} gives an example of a pseudo-linear distribution. Its ``accuracy-complexity plot" contains a pronounced local minimum at $FVE\approx 0.99$. Further increase of the number of nodes almost does not allow to increase accuracy. Thus, the optimal configuration of the approximator is achieved at 32 nodes and no branching.

A simple example of a branching data distribution is shown in Figure~\ref{star_tree} and Figure~\ref{star_curve}. In Figure~\ref{star_curve} we apply a reduced grammar  {\it ``Add a node to a terminal node"} which produces the principal curves (not trees). The complexity of the approximator in this case grows exponentially and suddenly saturates at $FVE\approx 0.971$ and 31 nodes. 16 nodes are needed to produce a principal curve grid approximation for the accuracy $FVE\approx 0.92$. By contrast, the full principal tree grammar {\it ``Add a node to any node, Bisect an edge"} needs only 6 nodes to achieve the same accuracy (Figure~\ref{star_tree}), and this corresponds to a local minimum of the complexity. Further increase of the accuracy does not lead to any significant increase of either geometrical or structural complexity. Figures~\ref{star_tree} and \ref{star_curve} illustrate dependence of the complexity measures on the grammar chosen: the correct principal tree grammar allows to construct much more optimal approximators.

Figure~\ref{tree_example} represents an interesting non-trivial example of a ``tree-like" structure with several complexity scales. The ``accuracy-complexity" plot contains two pronounced local minima: at $FVE\approx 0.87$ and $FVE\approx 0.98$. These two minima correspond to two scales of data approximation. The first scale depicts the data structure as a simple 3-star (corresponding to the barcode $1||4$). The second scale corresponds to the ``gestalt" formed by the data points: it is a combination of further branching, containing one 4-star and two 3-stars (the barcode is $1|2||14$). Further improvement of accuracy (after $FVE\approx 0.98$) is quite expensive in terms of geometrical complexity. The geometrical complexity increases by 4-fold and the number of nodes by 3.5 fold to gain only 1.5\% of the total variance explained.

\subsection{Examples from UC Irvine Machine Learning Repository}

\begin{figure}
 \subfigure
    {
        (a)~~\includegraphics[width=11.5cm]{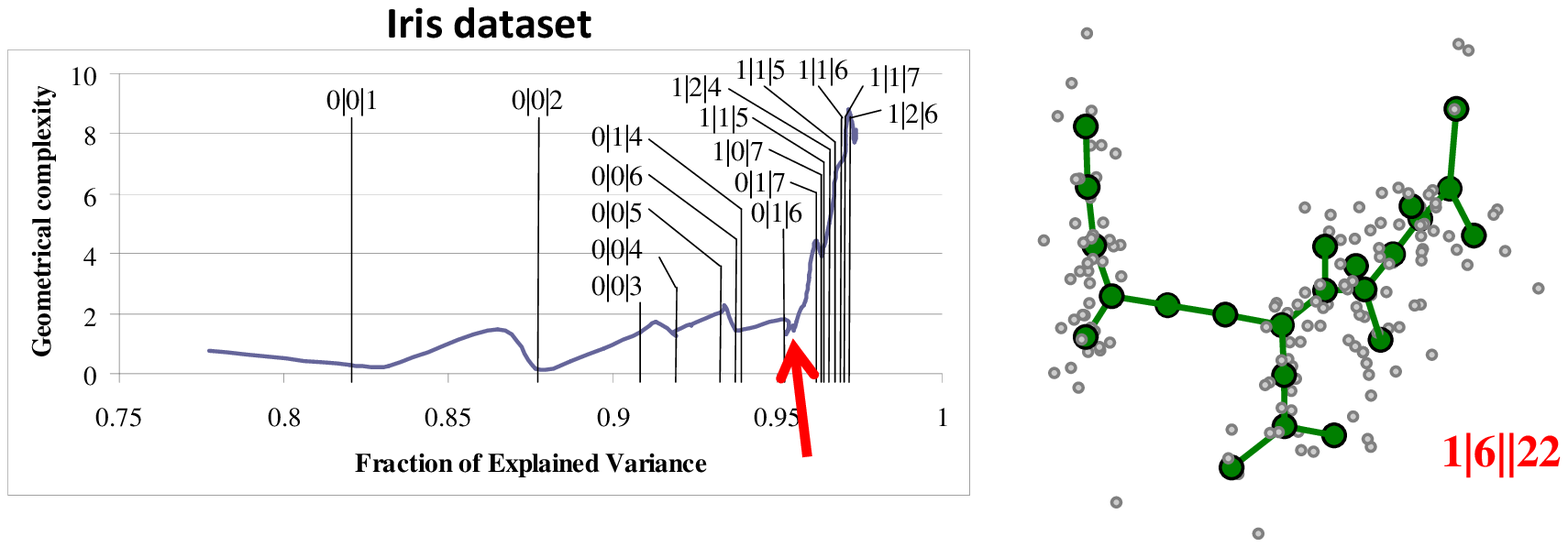}
        \label{iris}
    }
 \subfigure
    {
        (b)~~\includegraphics[width=11.5cm]{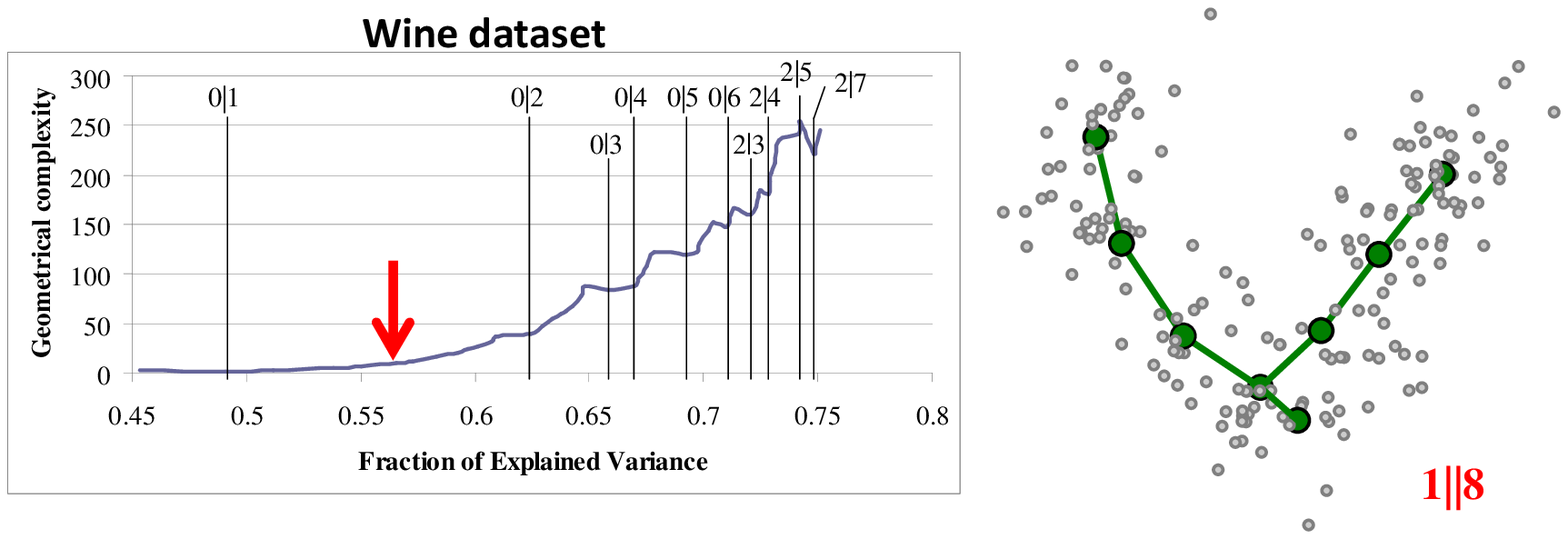}
        \label{wine}
    }
  \subfigure
    {
        (c)~~\includegraphics[width=11.5cm]{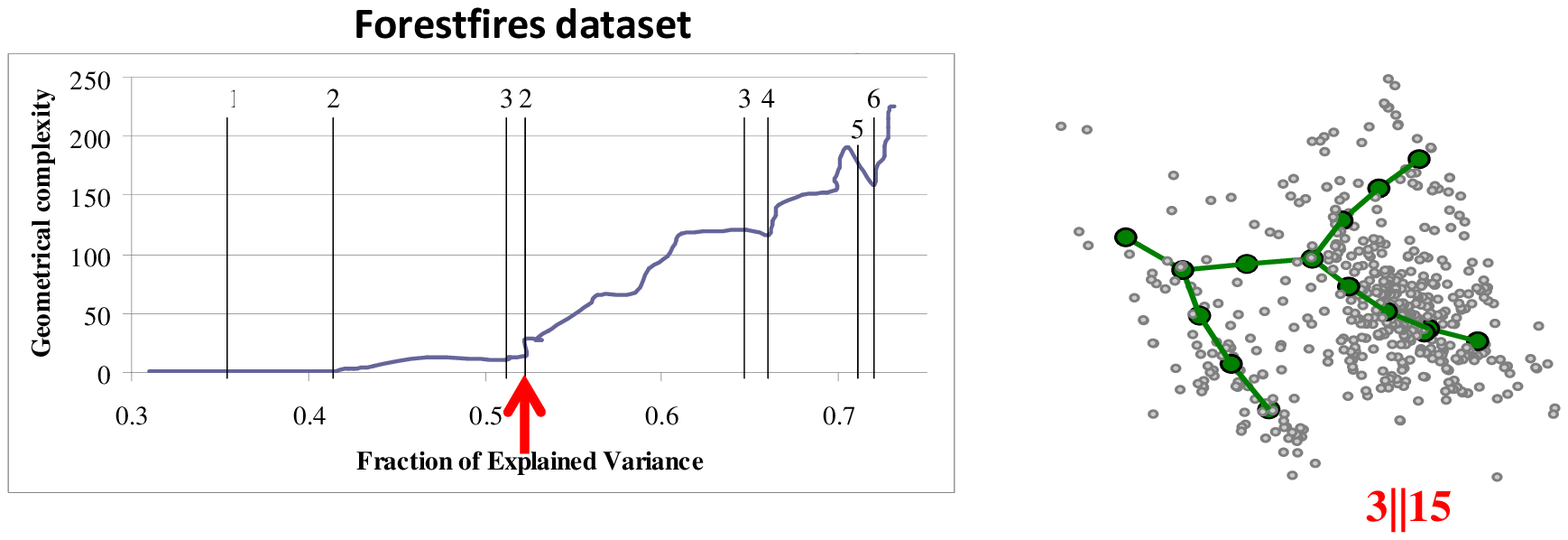}
        \label{forestfires}
    }
 \subfigure
    {
        (d)~~\includegraphics[width=11.5cm]{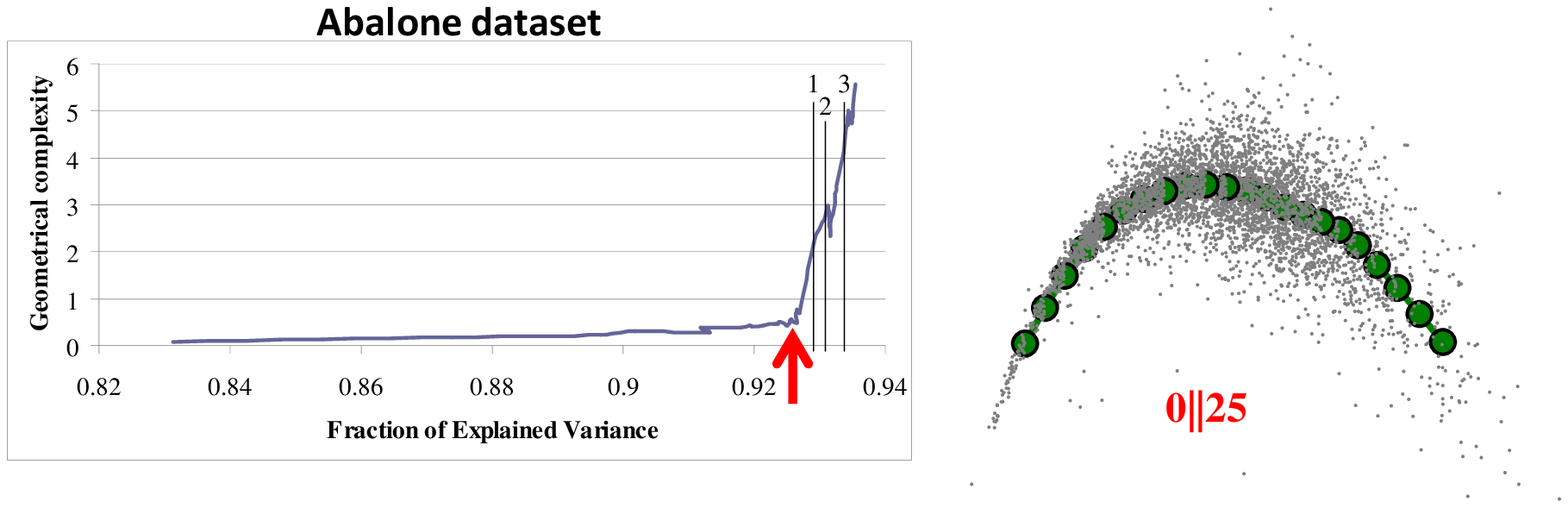}
        \label{abalone}
    }
\caption{\label{UCI_examples} Examples of the ``accuracy-complexity" graphs (on the left) for the real-life data distributions (on the right, shown in projection on the first two principal components). The arrows show the accuracy/complexity chosen to construct the principal tree shown on the right together with the data. The barcode shown together with the data distribution specifies the structural complexity. Part of the barcode (only the number of stars) is shown on the ``accuracy-complexity" plot above the vertical lines visualizing the discrete changes of the structural complexity.}
\end{figure}

We constructed principal trees for several datasets from the UCI Machine Learning Repository \cite{FrankAsuncion2010} and plotted the ``accuracy-complexity" graphs for them (Figure~\ref{UCI_examples}). In this Figure, both changes in the geometrical as well as structural complexity (using verrtical lines labeled by the structural complexity barcode) are visualized as a function of the approximator's accuracy.

The plots show significantly different complexity of the datasets which does not necessarily coincides with dimension of the dataset or the number of points in it. For example, the {\it Abalone} dataset with its 4177 points represents a simple pseudolinear distribution of points. Approximating it makes sense until the principal tree starts to form branches. When this happens, the complexity estimation abruptly goes up, meaning unnecessary growth of the approximator's complexity. Quite oppositely, the {\it Iris} dataset (150 data points) shows a non-trivial landscape of complexity, with many local minima and a constant growth of the structural complexity. The {\it Wine} dataset has a local minimum at only four nodes, forming a 3-star: this corresponds to existence of 3 well-separated ellipsoidally shaped clusters in the dataset. Further improvement of the approximation gradually and exponentially increases the approximator's complexity, and, after it is increased more than tenfold, the principal tree starts to branch further.

Finally, the {\it forestfires} dataset shows increase of accuracy and complexity in two epochs. During the first epoch,  the geometrical complexity practically does not grow. In the second epoch it increases approximately linearly with the accuracy and saturates at ($FVE \approx 0.72$). Interestingly, at the point of the epoch change ($FVE \approx 0.52$), the structural complexity fluctuates from ``$2 \vert \vert 11$"`to ``$3 \vert\vert 12$", back to ``$2\vert\vert 16$" and further to ``$3\vert\vert 27$", due to the trimming grammar applications (it becomes more advantageous to decrease the structural complexity during 11 iterations).

\section{Conclusion}

In the conclusion we should, first of all, repeat the principal guiding idea of this study: {\it good approximator is always characterized by a balance between the approximation accuracy and the approximator's complexity}. We define the data complexity as the complexity of its optimal approximator. Given the type of the approximator, one can estimate its complexity with respect to a dataset by looking at the ``accuracy-complexity" plot: the optimal approximator will correspond to such a point where the further increase of accuracy leads to the drastic increase of complexity. Often this corresponds also to a local minimum of the approximator's complexity. Several local minima of the complexity landscape correspond to several ``scales" of complexity in the data distribution, just as there exist multiple scales in estimating the intrinsic data dimensionality.

Good and flexible approximators allowing gradual increase of its complexity are principal cubic complexes, which can be constructed using a graph grammar. The simplest graph grammar {\it ``add a node; bisect an edge" } produce principal trees which can be used for measuring the data complexity and choosing an approximator with the most optimal accuracy/complexity ratio, corresponding to the selected complexity scale. We applied this method to several artificial and real-life datasets, and showed that the ``accuracy/complexity" plot contains enough information to justify such a choice.





\bibliographystyle{model1-num-names}
\bibliography{ZinovyevMirkes_CAMWA_2012}







\end{document}